\documentclass[10pt,twocolumn,letterpaper]{article}
\usepackage{iccv}

\usepackage{tikz}

\usepackage{times}
\usepackage{epsfig}
\usepackage{graphicx}
\usepackage{amsmath}
\usepackage{placeins}
\usepackage{hhline}

\newcommand{\fig}[1]{Figure~\ref{fig:#1}}
\newcommand{\sect}[1]{Section~\ref{sect:#1}}
\newcommand{\tab}[1]{Table~\ref{tab:#1}}

\newcommand{\eq}[1]{(\ref{eq:#1})}

\usepackage[breaklinks=true,bookmarks=false]{hyperref}

\iccvfinalcopy 

\def\iccvPaperID{0000} 

\ificcvfinal\fi

\begin{document}

\title{Escape from Cells: Deep Kd-Networks for the Recognition of\\ 3D Point Cloud Models}

\author{Roman Klokov\\
Skolkovo Institute of Science and Technology\\
{\tt\small roman.klokov@skoltech.ru}
\and
Victor Lempitsky\\
Skolkovo Insitute of Science and Technology\\
{\tt\small lempitsky@skoltech.ru}
}

\maketitle
\thispagestyle{empty}
\begin{abstract}
We present a new deep learning architecture (called Kd-network) that is designed for 3D model recognition tasks and works with unstructured point clouds. The new architecture performs multiplicative transformations and shares parameters of these transformations according to the subdivisions of the point clouds imposed onto them by kd-trees. Unlike the currently dominant convolutional architectures that usually require rasterization on uniform two-dimensional or three-dimensional grids, Kd-networks do not rely on such grids in any way and therefore avoid poor scaling behavior. In a series of experiments with popular shape recognition benchmarks, Kd-networks demonstrate competitive performance in a number of shape recognition tasks such as shape classification, shape retrieval and shape part segmentation.
\end{abstract}

\section{Introduction}

As the 3D world around us is getting scanned and digitized and as the archives of human-designed models are growing in size, recognition and analysis of 3D geometric models are gaining importance. Meanwhile, deep convolutional networks (ConvNets)~\cite{LeCun98} have excelled at solving analogous recognition tasks for 2D image datasets. It is therefore natural that a lot of research currently aims at the adaptation of deep ConvNets to 3D models \cite{Wu15,Maturana15,Brock16, Wu16,Wang15,Qi16a,Su15,Boscaini15,Boscaini16}.

Such adaptation is non-trivial. Indeed, the most straightforward way to make ConvNets applicable to 3D data, is to rasterize 3D models onto uniform voxel grids. Such approach however leads to excessively large memory footprints and slow processing times. Consequently, works that follow this path \cite{Wu15, Maturana15,Brock16, Wu16,Wang15,Li16} use small spatial resolutions (e.g.\ $64\times{}64\times{}64$), which clearly lag behind grid resolutions typical for processing 2D data, and is likely to be insufficient for the recognition tasks that require attention to fine details in the models.

To solve this problem, we take inspiration from the long history of research in computer graphics and computational geometry communities \cite{Samet90,Foley94}, where a large number of indexing structures that are far more scalable than uniform grids have been proposed, including kd-trees~\cite{Bentley75}, octrees~\cite{Meagher80}, binary spatial partition trees~\cite{Schumacker69}, R-trees~\cite{Guttman84}, constructive solid geometry~\cite{Requicha77}, etc. Our work was motivated by the question, whether at least \textit{some of these indexing structures are amenable for forming the base for deep architectures}, in the same way as uniform grids form the base for the computations, data alignment and parameter sharing inside convolutional networks.

In this work, we pick one of the most common 3D indexing structures (a kd-tree~\cite{Bentley75}) and design a deep architecture (a \textit{Kd-network}) that in many respects mimics ConvNets but uses kd-tree structure to form the computational graph, to share learnable parameters, and to compute a sequence of hierarchical representations in a feed-forward bottom-up fashion. In a series of experiments, we show that Kd-networks come close (or even exceed) ConvNets in terms of accuracy for recognition operations such as classification, retrieval and part segmentation. At the same time, Kd-networks come with smaller memory footprints and more efficient computations at train and at test time thanks to the improved ability of kd-trees to index and structure 3D data as compared to uniform voxel grids.

Below, we first review the related work on convolutional networks for 3D models in \sect{related}. We then discuss the Kd-network architecture in \sect{method}. An extensive evaluation on toy data (a variation of MNIST) and standard benchmarks (ModelNet10, ModelNet40, SHREC'16, ShapeNet part datasets) is presented in \sect{experiments}. We summarize the work in \sect{conclusion}.

\section{Related Work}
\label{sect:related}

Several groups investigated application of ConvNets to the rasterizations of 3D models on uniform 3D grids \cite{Wu15, Maturana15}. The improvements include combinations of generative and very deep discriminative architectures \cite{Brock16, Wu16}. Despite considerable success in coarse-level classification, the reliance on uniform 3D grids for data representation makes scaling of such approaches to fine-grained tasks and high spatial representations problematic. To improve the scalability \cite{Wang15,Li16} have considered sparse ways to define convolutions, while still using uniform 3D grids for representations.

Another approach~\cite{Su15,Qi16a} is to avoid the use of 3D grids, and instead apply two-dimensional ConvNets to 2D projections of 3D objects, while pooling representations corresponding to different views. Despite gains in efficiency, such approach may not be optimal for hard 3D shape recognition tasks due to the loss of information associated with the projection operation. A group of approaches (such as spectral ConvNets~\cite{Bruna13,Boscaini15} and anisotropic ConvNets~\cite{Boscaini16}) generalize ConvNets to non-Euclidean geometries, such as mesh surfaces. These have shown very good performance for local correspondence/matching tasks, though their performance on standard shape recognition and retrieval benchmarks has not been reported. Kd-networks as well as the PointNet architecture \cite{Qi16b} work directly with points and therefore can take the representations computed with intrinsic ConvNets as inputs. Such configuration is likely to combine at least some of the advantages of extrinsic and intrinsic ConvNets, but its investigation is left for future work.

Aside from their connections to convolutional networks that we discuss in detail below, Kd-networks are related to recursive neural networks \cite{Socher11}. Both recursive neural networks and Kd-networks have tree-structured computational graphs. However, the former share parameters across all nodes in the computational tree graph, while sharing of parameters in Kd-networks is more structured, which allows them to achieve competitive performance.

Finally, two approaches developed in parallel to ours share important similarities. OctNets~\cite{Riegler17} are modified ConvNets that operate on non-uniform grids (shallow OctTrees) and thus share the same idea of utilizing non-uniform spatial structures within deep architectures. Even more related are graph-based ConvNets with edge-dependant filters \cite{Simonovsky17}. Kd-networks can be regarded as a particular instance of their architecture with a kd-tree being an underlying graph (whereas \cite{Simonovsky17} evaluated nearest neighbor graphs for point cloud classification). Kd-networks outperform both \cite{Riegler17} and the setup in \cite{Simonovsky17} on the ModelNet benchmarks suggesting that deep architectures based on kd-trees may be particularly well suited for coarse-level shape categorization.

\newcommand{\x}{\mathtt{x}}
\newcommand{\y}{\mathtt{y}}
\newcommand{\z}{\mathtt{z}}
\newcommand{\tk}{\mathbf{t}}
\newcommand{\Sk}{S}
\renewcommand{\v}{\mathbf{v}}
\newcommand{\vt}{\tilde{\mathbf{v}}}
\newcommand{\f}{\mathbf{f}}
\newcommand{\W}[2]{W_{#2}^{#1}}
\newcommand{\Wt}[2]{\tilde{W}_{#2}^{#1}}
\newcommand{\Wx}[1]{W_\x^{#1}}
\newcommand{\Wy}[1]{W_\y^{#1}}
\newcommand{\Wz}[1]{W_\z^{#1}}
\newcommand{\bb}[2]{\mathbf{b}_{#2}^{#1}}
\newcommand{\bbt}[2]{\tilde{\mathbf{b}}_{#2}^{#1}}
\newcommand{\bx}[1]{\mathbf{b}_\x^{#1}}
\newcommand{\by}[1]{\mathbf{b}_\y^{#1}}
\newcommand{\bz}[1]{\mathbf{b}_\z^{#1}}
\newcommand{\T}{\mathcal{T}}

\section{Shape Recognition with Kd-Networks}
\label{sect:method}

\begin{figure}
    \centering
    \scalebox{0.9}{
    \includegraphics[width=\columnwidth]{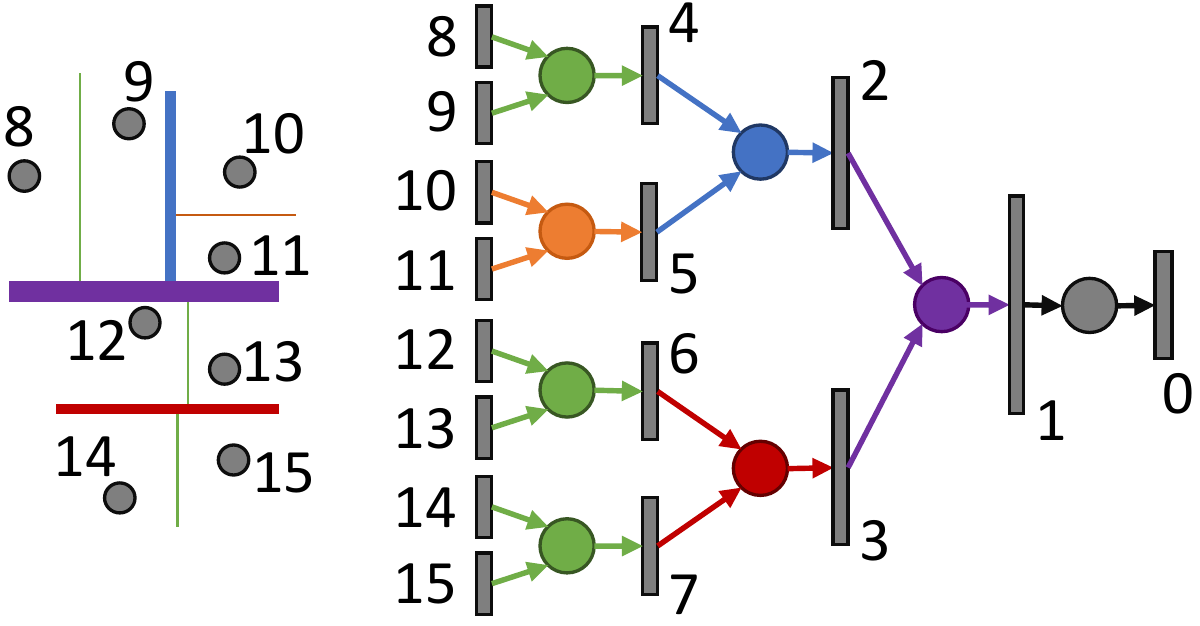}
    }
    \caption{A kd-tree built on the point cloud of eight points (left), and the associated Kd-network built for classification (right). We number nodes in the kd-tree from the root to leaves. The arrows indicate information flow during forward pass (inference). The leftmost bars correspond to leaf (point) representations. The rightmost bar corresponds to inferred class posteriors $\v_0$.  Circles correspond to affine transformations with learnable parameters. Colors of the circles indicate parameter sharing, as splits of the same type (same orientation, same tree level -- three ``green'' splits in this example) share the transformation parameters.}
    \label{fig:kdnet}
\end{figure}

We now introduce Kd-networks, starting with the discussion of their input format (kd-trees of certain size), then discussing the bottom-up computation of representations performed by Kd-networks, and finally discussing supervised parameter learning.

\subsection{Input}
The new deep architecture (the Kd-network) works with kd-trees constructed for 3D point clouds. Kd-networks can also consider and utilize properties of individual input points (such as color, reflectivity, normal direction) if they are known. At train time, Kd-network works with point clouds of a fixed size $N = 2^D$ (point clouds of different sizes can be reduced to this size using sub- or oversampling). A kd-tree is constructed recursively in a top-down fashion by picking the coordinate axis with the largest range (span) of point coordinates, and splitting the set of points into two equally-sized subsets, subsequently recursing to each of them. As a result, a balanced kd-tree $\T$ of depth $D$ is produced that contains $N{-}1 = 2^D{-}1$ non-leaf nodes. 

Each non-leaf node $V_i \in \T$ is thus associated with one of three splitting directions $d_i$ (along $x$, $y$ or $z$-axis, i.e.\ $d_i\in \{\x, \y, \z\}$) and a certain split position (threshold) $\tau_i$. A tree node is also characterized by the level $l_i \in \{1,..,D-1\}$, with $l_i{=}1$ for the root node, and $l_i{=}D$ for tree leaves that contain individual 3D points. We assume that the nodes in the balanced tree are numbered in the standard top-down fashion, with the root being the first node, and with the $i$th node having children with numbers $c_1(i)=2i$ and $c_2(i)=2i+1$.

\subsection{Processing data with Kd-networks}
Given an input kd-tree $\T$, a pretrained Kd-network computes vectorial representations $\v_i$ associated with each node of the tree. For the leaf nodes these representations are given as $k$-dimensional vectors describing the individual points, associated with those leaves. The representations corresponding to non-leaf nodes are computed in the bottom-up fashion (\fig{kdnet}). Consider a non-leaf node $i$ at the level $l(i)$ with children $c_1(i)$ and $c_2(i)$ at the level $l(i)+1$, for which the representations $\v_{c_1(i)}$ and $\v_{c_2(i)}$ have already been computed. Then, the vector representation $\v_i$ is computed as follows:
\begin{equation} \label{eq:main}
    \v_i = \begin{cases} \phi( \Wx{l_i} [\v_{c_1(i)}; \v_{c_2(i)}] + \bx{l_i}), \text{ if } d_i = \x\,,\\
     \phi( \Wy{l_i} [\v_{c_1(i)}; \v_{c_2(i)}] + \by{l_i}), \text{ if } d_i = \y\,,\\
      \phi( \Wz{l_i} [\v_{c_1(i)}; \v_{c_2(i)}] + \bz{l_i}), \text{ if } d_i = \z\,,
      \end{cases}
\end{equation}
or in short form:
\begin{equation}  \label{eq:main2}
    \v_i =  \phi( \W{l_i}{d_i} [\v_{c_1(i)}; \v_{c_2(i)}] + \bb{l_i}{d_i})\,.  
\end{equation}
Here, $\phi(\cdot)$ is some non-linearity (e.g.\ REctified Linear Unit $\phi(a) = \max(a,0)$), and square brackets denote concatenation. The affine transformation in \eq{main} is defined by the learnable parameters $\{\Wx{l_i},\Wy{l_i},\Wz{l_i},\bx{l_i},\by{l_i},\bz{l_i}\}$ of the layer $l_i$. Thus, depending on the splitting direction $d_i$ of the node, one of the three affine transformations followed by a simple non-linearity is applied. 

The dimensionality of the matrices and the bias vectors are determined by the dimensionalities $m^1,m^2,\dots,m^D$ of representations at each level of the tree. The $\Wx{l}$ ,$\Wy{l}$, and $\Wz{l}$ matrices at the $l$th level thus have the dimensionality $m^l{\times}2m^{l+1}$ (recall that the levels are numbered from the root to the leaves) and the bias vectors $\bx{l},\by{l},\bz{l}$ have the dimensionality $m^l$.

Once the transformations \eq{main} are applied in a bottom-up order, the root representation $\v_1(\T)$ for the sample $\T$ is obtained. Naturally, it can be passed through several additional linear and non-linear transformations (``fully-connected layers''). In our classification experiments, we directly learn linear classifiers using $\v_1(\T)$ representation as an input. In this case, the classification network output the vector of unnormalized class odds:
\begin{equation} \label{eq:classifier}
    \v_0(\T) = W^0 \v_1(\T) + \mathbf{b}^0\,,
\end{equation}
where $W^0$ and $\mathbf{b}^0$ are the parameters of the final linear multi-class classifier.

\begin{figure}
    \centering
    \scalebox{0.9}{
    \includegraphics[width=\columnwidth]{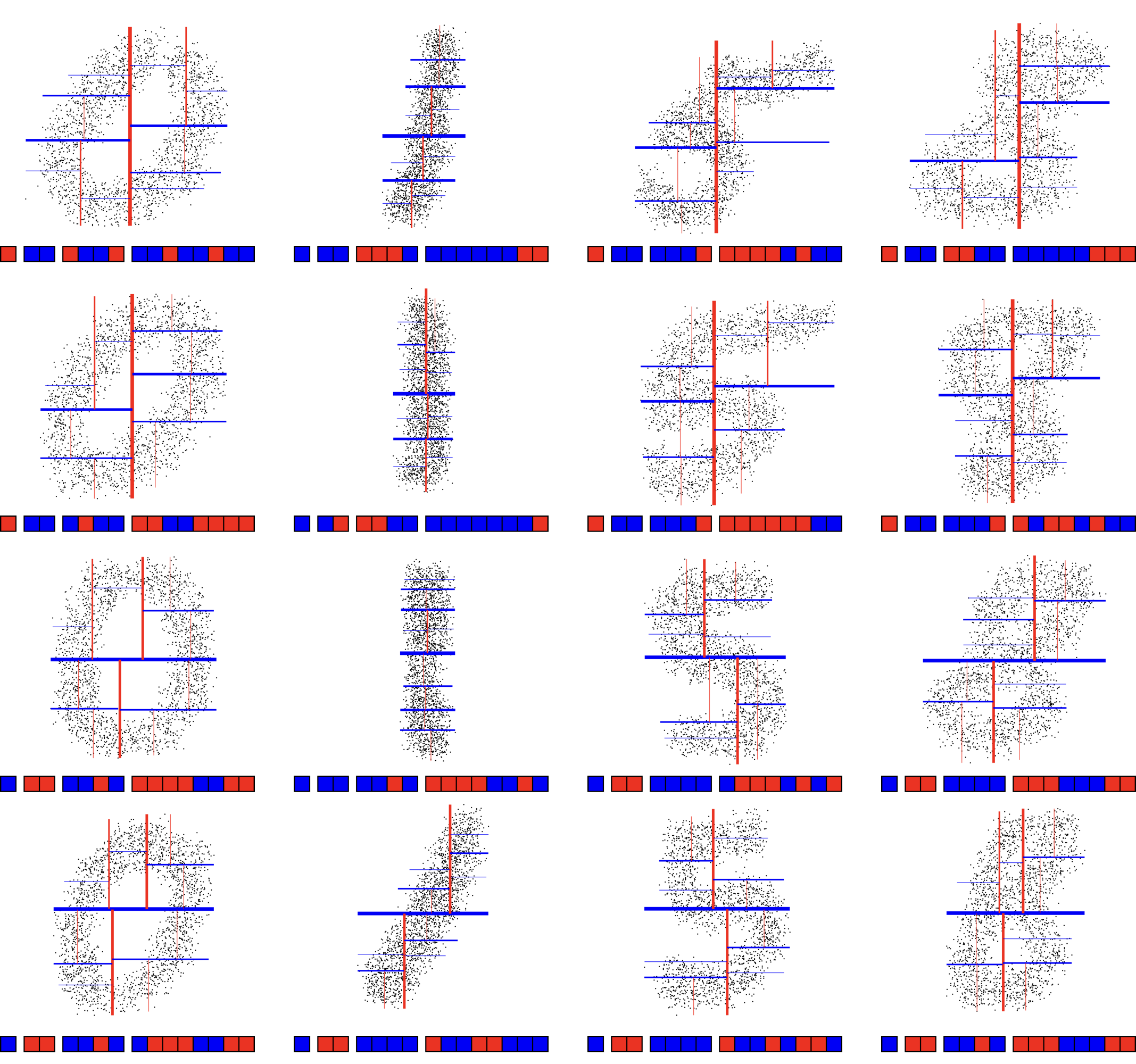}
    }
    \caption{Kd-trees for MNIST clouds. We visualize several examples of 2D point clouds for MNIST (see text for description) with constructed kd-trees. The type of split is encoded with color and for each example the types of splits for the first four levels of the tree are shown below. Importantly, the structure of the kd-tree serves as a shape descriptor (e.g.\ `ones' are dominated by vertical splits, and `zeroes' tend to interleave vertical and horizontal splits as a kd-tree is traversed from the root to a leaf).}
    \label{fig:kd_trees_mnist}
\end{figure}

\subsection{Learning to classify}
A Kd-network is a feed-forward neural network that has the learnable parameters $\{\Wx{j},\Wy{j},\Wz{j},\bx{j},\by{j},\bz{j}\}$ at each of the $D{-}1$ non-leaf levels $j \in \{1..D{-}1\}$, as well as the learnable parameters $\{W^0, \mathbf{b}^0\}$ for the final classifier. Standard backpropagation method can be used to compute the gradient of the loss function w.r.t.\ network parameters. The network parameters can thus be learned from the dataset of labeled kd-trees using standard stochastic optimization algorithms and standard losses, such as cross-entropy on the network outputs $\v_0(\T)$ \eq{classifier}.

\subsection{Learning to retrieve}
It is straightforward to learn the representation \eq{classifier} to produce not the class odds, but a descriptor vector of a certain dimensionality that characterizes the shape and can be used for retrieval. The parameters of the Kd-network can then be learned using backpropagation using any of the embedding-learning losses that observe examples of matching (e.g.\ same-class) and non-matching (e.g.\ different-class) shapes. In our experiments, we use a recently proposed histogram loss \cite{Ustinova16}, but more traditional losses such as Siamese loss \cite{Bromley93,Chopra05} or triplet loss \cite{Schultz04} could be used as well.

\subsection{Properties of Kd-networks}
Here we discuss the properties of the Kd-networks and also relate them to some of the properties of ConvNets.

\textbf{Layerwise parameter sharing.} Similarly to ConvNets, Kd-networks process the inputs by applying a sequence of parallel spatially-localized multiplicative operations interleaved with non-linearities. Importantly, just as ConvNets share their parameters for localized multiplications (convolution kernels) across different spatial locations, Kd-networks also share the multiplicative parameters $\{\Wx{j},\Wy{j},\Wz{j},\bx{j},\by{j},\bz{j}\}$ across all nodes at the tree level $j$.

\textbf{Hierarchical representations.} ConvNets apply bottom-up processing and compute a sequence of representations that correspond to progressively large parts of images. The procedure is hierarchical, in the sense that a representation of a spatial location at a certain layer is obtained from the representations of multiple surrounding locations at the preceding layer using linear and non-linear operations. All this is mimicked in Kd-networks, the only difference being that the receptive fields of two different nodes at the same level of the kd-tree are non-overlapping.

\textbf{Partial invariance to jitter.} Convolutional networks that use pooling operations and/or strides larger than one are known to possess partial invariance to small spatial jitter in the input. Kd-networks are also invariant to such jitter (unless such jitter strongly perturbs the representations of leaf nodes). This is because the key forward-propagation operation \eq{main} does ignore splitting thresholds $\tau_i$. Thus, any small spatial perturbation of input points that leave the topology of the kd-tree intact can only affect the output of a Kd-network via the leaf representations (which as will be revealed in the experiments play only secondary role in kd-networks).

\textbf{Non-invariance to rotations.} Similarly to ConvNets, Kd-networks are not invariant to rotations, as the underlying kd-trees are not invariant to them. In this aspect, Kd-networks are inferior to intrinsic ConvNets~\cite{Bruna13,Boscaini15,Boscaini16}. Standard tricks to handle variable orientations include pre-alignment (using heuristics or network branches that predict geometric transformations of the data \cite{Jaderberg15,Qi16b}) as well as pooling over augmentations~\cite{Laptev16} (or simply training with excessive augmentations).

\textbf{Role of kd-tree structure.} The role of the underlying kd-trees in the process of Kd-network data processing is two-fold. Firstly, the underlying kd-tree determines which leaf representations are getting combined/merged together and in which order. Secondly, the structure of the underlying kd-tree can be regarded as a shape descriptor itself (\fig{kd_trees_mnist}) and thus serves as the source of the information irrespective of what the leaf representations are. The Kd-network then serves as a mechanism for extracting the shape information contained in the kd-tree structure. As will be revealed in the experiments, the second aspect is of considerable importance, as even in the absence of meaningful leaf representations, Kd-networks are able to recognize shapes well solely based on the kd-tree structure.

\subsection{Extension for segmentation}
\label{sect:segm}

\begin{figure}
    \centering
    \includegraphics[width=\columnwidth]{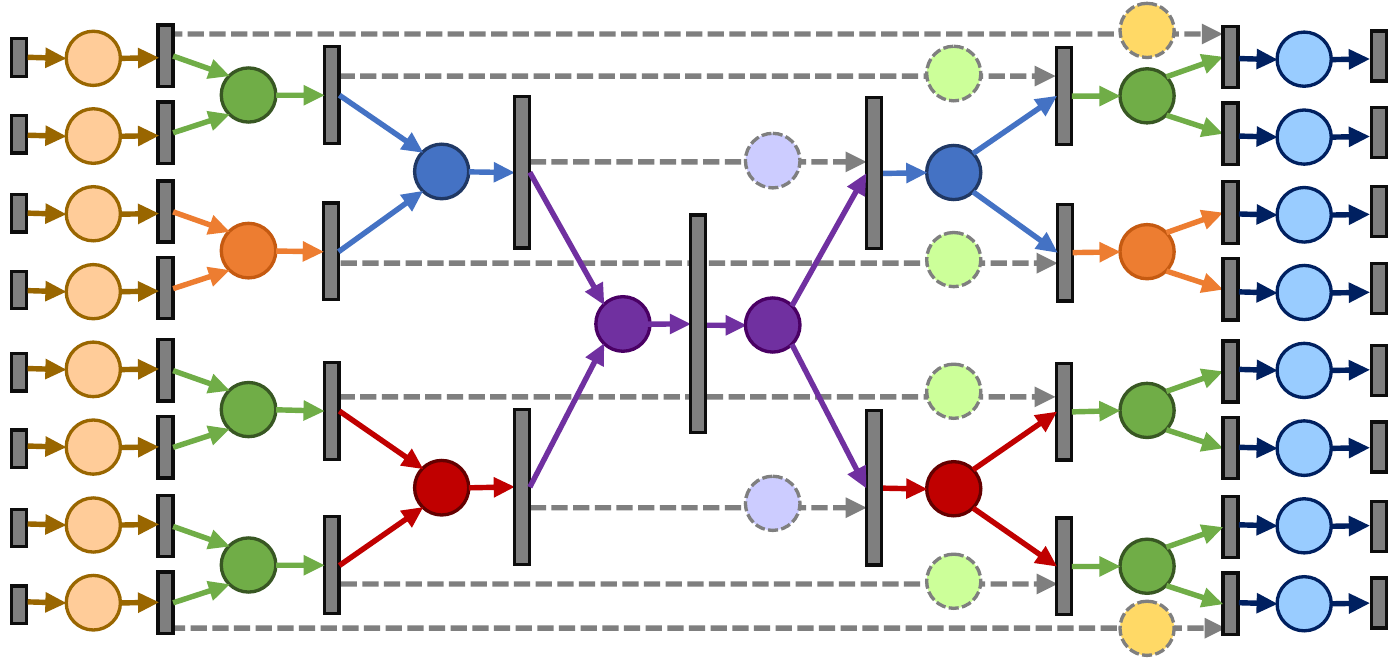}
    \caption{The architecture for parts segmentation (individual point classification) for the point cloud shown in \fig{kdnet} (left). Arrows indicate computations that transform the representations (bars) of different nodes. Circles correspond to affine transformations followed by non-linearities. Similarly colored circles on top of each other share parameters. Dashed lines correspond to skip-connections (some ``yellow'' skip connections are not shown for clarity). The input representations are processed by an additional transformation (light-brown) and there are additional transformations applied to every leaf representation independently at the end of the architecture (light-blue).}
    \label{fig:kdsegm}
\end{figure}

Kd-network architecture can be extended to perform semantic/part segmentation tasks in the same way as ConvNets. In this work, we mimic the encoder-decoder (hourglass-shaped) architecture with \textit{skip connections} (\fig{kdsegm}) that has been proposed for ConvNets in \cite{Long15,Ronneberger15}. More formally, during inference firstly the representations $\v_i$ are computed using \eq{main2}, and then the second representation vector $\vt_i$ is computed at each node $i$. The computations of the second representation proceed by setting $\vt_1{=}\v_1$ (or obtaining $\vt_1$ by one or several fully connected layers) and then using the following chain of top-down computations:
\begin{align}
  &\vt_{c_1(i)}=\phi([\Wt{l_i}{d_{c_1(i)}}\vt_i+\bbt{l_i}{d_{c_1(i)}};\Sk^{l_i}\v_{c_1(i)}+\tk^{l_i}])\,,\nonumber\\ 
  &\vt_{c_2(i)}=\phi([\Wt{l_i}{d_{c_2(i)}}\vt_i+\bbt{l_i}{d_{c_2(i)}};\Sk^{l_i}\v_{c_2(i)}+\tk^{l_i}])\,, \label{eq:up}
\end{align}
where $\Wt{l_i}{d_{c_*(i)}}$ and $\bbt{l_i}{d_{c_*(i)}}$ are the parameters of the affine transformation that map the parent's representation to the children representations stacked on top of each other, while $\Sk^{l_i}$ and $\tk^{l_i}$ are the parameters of the affine transformation within the skip connection from $\v_{c_1(i)}$ to $\vt_{c_1(i)}$ (as well as from $\v_{c_2(i)}$ to $\vt_{c_2(i)}$). In our implementation, the former set of parameters depends on split orientation, while the latter depends on the node layer only.

To increase the capacity of the model, additional multiplicative layers interleaved with non-linearities can be inserted in the beginning of the architecture or at the end of architecture (with parameters shared across leaves making these layers analogous to $1{\times}1$-convolutions in ConvNets). Also, fully-connected multiplicative layers can be inserted at the bottleneck. 

\subsection{Implementation details}

\textbf{Leaf representation.} As mentioned above, for a leaf node $i$ a representation $\v_i$ can be defined in several ways. In our experiments, unless stated otherwise, we use normalized 3D coordinates obtained by putting the center of mass of the shape at origin and rescaling the input point cloud to fit the $[-1;1]^3$ 3D box. 

\textbf{Data augmentation.} Similarly to other machine learning architectures, performance of Kd-networks can be improved through training data augmentations. Below, we experiment with applying perturbing geometric transformations to 3D point clouds. Additionally, we found the injecting randomness into kd-tree construction very useful. For that, we randomize the choice of split directions using the following probabilities:
\begin{equation} \label{eq:distribution}
    P\left(d_i = j|\hat{\bf r}_i\right) = \frac{\exp{\gamma \hat{r}_i^j}}{\sum_{j=x, y, z}\exp{\gamma \hat{r}_i^j}}\,,
\end{equation}
where $\hat{r}_i$ is a vector of ranges normalized to unit sum.

\begin{table}
  \begin{center}
  \tabcolsep=0.11cm
\scalebox{0.95}{
\begin{tabular}{|l|c|c|c|c|}
\hline
\multicolumn{1}{|c}{\small ModelNet} & \multicolumn{2}{|c|}{\small 10-class}                             & \multicolumn{2}{c|}{\small 40-class}\\
\hline
\multicolumn{1}{|c|}{\small Accuracy averaging} & \multicolumn{1}{l|}{\small class} & \multicolumn{1}{l|}{\small instance} & \multicolumn{1}{l|}{\small class} & \multicolumn{1}{l|}{\small instance} \\
\hline
{\small 3DShapeNets \cite{Wu15}}     & {\small 83.5} & {\small -}    &                                                    {\small 77.3} & {\small -}\\
{\small MVCNN \cite{Su15}}           & {\small -}    & {\small -}    &                                                    {\small 90.1} & {\small -}\\
{\small FusionNet \cite{Hegde16}}    & {\small -}    & {\small 93.1} &                                                    {\small -}    & {\small 90.8}\\
{\small VRN Single \cite{Brock16}}   & {\small -}    & {\small 93.6} &                                                    {\small -}    & {\small 91.3}\\
{\small MVCNN \cite{Qi16a}}          & {\small -}    & {\small -}    &                                                    {\small 89.7} & {\small 92.0}\\
{\small PointNet \cite{Qi16b}}       & {\small -}    & {\small -}    &                                                    {\small 86.2} & {\small 89.2}\\
{\small OctNet \cite{Riegler17}}     & {\small 90.1} & {\small 90.9} &                                                    {\small 83.8} & {\small 86.5}\\
{\small ECC \cite{Simonovsky17}}     & {\small 90.0} & {\small 90.8} &                                                    {\small 83.2} & {\small 87.4}\\
\hline
{\small Kd-Net (depth 10)}           & {\small 92.8} & {\small 93.3} &                                                    {\small 86.3} & {\small 90.6}\\
{\small Kd-Net (depth 15)}           & {\small 93.5} & {\small 94.0} &                                                    {\small 88.5} & {\small 91.8}\\
\hline
{\small VRN Ensemble \cite{Brock16}} & {\small -}    & {\small 97.1} &                                                    {\small -}    & {\small 95.5}\\
{\small MVCNN-MultiRes \cite{Qi16a}} & {\small -}    & {\small -}    &                                                    {\small 91.4} & {\small 93.8}\\
\hline
\end{tabular}}
  \end{center}
  \caption{Classification results on ModelNet benchmarks. Comparison of accuracies of Kd-networks (depth 10 and 15) with state-of-the-art. Kd-networks outperform all single model architectures except MVCNNs, while performing worse than reported ensembles.}
  \label{tab:class_ext}
\end{table}

\section{Experiments}
\label{sect:experiments}

We now discuss the results of application of Kd-networks to shape classification, shape retrieval and part segmentation tasks benchmarks. For classification, we also evaluate several variations and ablations of Kd-networks. Our implementation of Kd-networks using Theano~\cite{Theano16} and Lasagne~\cite{Dieleman15} as well as additional qualitative and quantitative results are available at project webpage\footnote{http://sites.skoltech.ru/compvision/kdnets/}.

\subsection{Shape classification}

\textbf{Datasets and data processing.} We evaluate Kd-networks on datasets of 2D (for illustration purposes) as well as 3D point clouds. 2D point clouds were produced from the \textbf{MNIST} dataset~\cite{LeCun98} by turning centers of non-zero pixels into 2D points. A point cloud of a needed size was then sampled from the resulting set of points with an addition of a small random noise. \fig{kd_trees_mnist} shows examples of resulting point clouds.

\begin{table}
  \begin{center}
  \tabcolsep=0.11cm
\scalebox{0.95}{
\begin{tabular}{|l|c|c|c|}
\hline
            & {\small MNIST} & {\small ModelNet10} & {\small ModelNet40}\\
\hline
{\small Split-based linear}       & {\small 82.4} & {\small 83.4} & {\small 73.2}\\
{\small Kd-net RT{+}SA (no leaf)} & {\small 98.6} & {\small 92.7} & {\small 89.8}\\
\hline
{\small Kd-net DT}                & {\small 98.9} & {\small 89.2} & {\small 85.7}\\
{\small Kd-net RT}                & {\small 99.1} & {\small 92.8} & {\small 89.9}\\
{\small Kd-net RT{+}TA}           & {\small 99.1} & {\small 92.9} & {\small 90.1}\\
{\small Kd-net RT{+}SA}           & {\small 99.1} & {\small 93.2} & {\small 90.6}\\
{\small Kd-net RT{+}SA{+}TA}      & {\small 99.1} & {\small 93.3} & {\small 90.6}\\
\hline
\end{tabular}}
  \end{center}
  \caption{Classification accuracy for baselines and different data augmentations. The resulting accuracies for the baseline model, the ablated model with trivial leaf representations, as well as Kd-networks trained with various data augmentations. DT = deterministic kd-trees, RT = randomized kd-trees, TA = translation augmentation, SA = anisotropic scaling augmentation. All networks are depth 10. See text for discussion. }
  \label{tab:class_int}
\end{table}

The 10-class and the 40-class variations of ModelNet~\cite{Wu15} (\textbf{ModelNet10} and \textbf{ModelNet40}) benchmarks, containing 4899 and 12311 models respectively, were used for 3D shape classifications. The two datasets are split into the training set (3991 and 9843 models) and the test set (909 and 2468 models respectively). In this case, 3D point clouds were computed as follows: firstly, a given number of faces were sampled with the probability proportionate to their surface areas. Then, for the sampled face a random point was taken. The whole sampling procedure thus closely approximated uniform sampling of model surfaces.

\textbf{Training and test procedures.} Additionally we preprocess each object by applying a geometric perturbation and noise (as discussed below). Either a deterministic or a randomized kd-tree is constructed and, finally, the resulting point cloud and leaf representations are used to perform forward-backward pass in the Kd-Network. At test time, we use the same augmentations as were used during training and average predicted class probabilities over ten runs.

We experimented with the following augmentations: (i) proportional translations along every axis (\textit{TR}) of up to $\pm{}0.1$ in normalized coordinates; proportional anisotropic rescaling over the two horizontal axes (\textit{AS}) by the number sampled from the $0.66$ to $1.5$ range. More global augmentations like flips or rotations did not improve results. Additionally, we evaluated both deterministic (DT) and randomized (RT) kd-trees. For our experiments we fixed the parameter $\gamma$ in \eq{distribution} to ten.

\textbf{Benchmarking classification performance.} We compare our approach to the state-of-the-art on the ModelNet10 and ModelNet40 benchmarks in \tab{class_ext}. We give the results obtained with kd-trees of depth 10 and depth 15. For depth 10, our architecture firstly obtains leaf representation of size 32 from initial points coordinates with an affine transformation with parameters shared across all the input points interleaved with a ReLU non-linearity, then a \textit{Kd-network} obtains intermediate representations of sizes: $32-64-64-128-128-256-256-512-512-128$. Resulting representation for a point cloud is directly used to obtain class posteriors with a single fully connected layer. For depth 15, the previous architecture has been modified by changing the size of leaf representation to 8 and by updated progression of intermediate representation sizes: $16-16-32-32-64-64-128-128-256-256-512-512-1024-1024-128$. 

In both cases, we used translation-based and anisotropic scaling-based augmentations as well as randomized kd-tree generation at test and at train time. Note that despite the use of random augmentations, a single model (i.e.\ a single set of model weigths) was evaluated for each of the cases (depth 10 and depth 15). Our results are better than all previous single-model results on these benchmarks except MVCNNs. While being worse than the reported ensembles, Kd-networks can be trained faster. VRN ensemble involves 6 models each trained over the course of 6 days on NVidia Titan X. Our depth-10 model can be trained in 16 hours, and our depth-15 model can be trained in 5 days using an older NVidia Titan Black. Furthermore, more than 75\% of the time is spent on point cloud sampling and kd-tree fitting, while the training itself takes less then a quarter of the mentioned times.  

It is also interesting to note that the performance of Kd-networks on the MNIST dataset reaches 99.1\% (\tab{class_int}), which is in the ballpark of the results obtained with ConvNets (without additional tricks).

\textbf{Ablations and variants.} Kd-networks use two sources of information about each object, namely the leaf representations and the direction of the splits. Note, that the split coordinates are not used in the classification. We assess the relative importance of the two sources of the information using two baselines. Firstly, we consider the baseline for both 2D and 3D point clouds that encode split information from their kd-trees in the following way: every split on every level is one-hot encoded and concatenated to resulting feature vector. We then use a linear classifier on such a representation (which is also shown as red/blue bars in \fig{kd_trees_mnist}). This baseline evaluates how much information can be recovered from the split orientation information with very little effort.

We also evaluate a model ablation corresponding to our full method with the exception that we remove the first source information. To this end, we make each leaf representation equal a one-dimensional vector (i.e.\ scalar) that equals one, effectively removing the first source of information.

The results in \tab{class_int} suggest that the first (linear classification) baseline performs much worse than Kd-network (even without leaf information), which suggests that multi-stage hierarchical data flow and intricate weight sharing mechanism of Kd-networks plays an important role (note, however that this baseline performs considerably better than chance suggesting that the orientation of splits in a kd-tree can serve as shape descriptor). Most interestingly, the ablated version of Kd-network comes very close to the full method, highlighting that the second source of information (split direction) dominates the first in terms of importance (confirming the suitability of kd-trees for shape description). 

\begin{table*}
  \begin{center}
  \tabcolsep=0.11cm
\scalebox{0.95}{
\begin{tabular}{|l|ccccc|ccccc|}
    \hline
        & \multicolumn{5}{c|}{Micro}                                                                                         & \multicolumn{5}{c|}{Macro}                                                                                         \\
    \hline
        & P@N   & \multicolumn{1}{l}{\small R@N} & \multicolumn{1}{l}{\small F1@N} & \multicolumn{1}{l}{\small mAP} & \multicolumn{1}{l|}{\small NDCG@N} & {\small P@N}   & \multicolumn{1}{l}{\small R@N} & \multicolumn{1}{l}{\small F1@N} & \multicolumn{1}{l}{\small mAP} & \multicolumn{1}{l|}{\small NDCG@N} \\
    \hline
Bai~\cite{Savva16}     & 0.706 & 0.695 & 0.689 & 0.825 & 0.896 & 0.444 & 0.531 & 0.454 & 0.740 & 0.850 \\
Su~\cite{Savva16}      & 0.770 & 0.770 & 0.764 & 0.873 & 0.899 & 0.571 & 0.625 & 0.575 & 0.817 & 0.880 \\
Kd-net (depth 15)      & 0.760 & 0.768 & 0.743 & 0.850 & 0.905 & 0.492 & 0.676 & 0.519 & 0.746 & 0.864 \\
    \hhline{|=|=====|=====|}
Bai~\cite{Savva16}     & 0.678 & 0.667 & 0.661 & 0.811 & 0.889 & 0.414 & 0.496 & 0.423 & 0.730 & 0.843 \\
Su~\cite{Savva16}      & 0.632 & 0.613 & 0.612 & 0.734 & 0.843 & 0.405 & 0.484 & 0.416 & 0.662 & 0.793 \\
Kd-net (depth 15)      & 0.473 & 0.519 & 0.451 & 0.617 & 0.814 & 0.205 & 0.529 & 0.241 & 0.484 & 0.726 \\
MVKd-net (depth 10)    & 0.660 & 0.652 & 0.631 & 0.766 & 0.868 & 0.355 & 0.560 & 0.382 & 0.617 & 0.792 \\
    \hline
\end{tabular}}
  \end{center}
  \caption{Retrieval results on normal and perturbed (top and bottom respectively) version of ShapeNetCore dataset for the metrics introduced in \cite{Savva16} (higher is better). See \cite{Savva16} for the details of metric and the presented systems (in general, all systems in \cite{Savva16} incorporated some variants of 2D multi-view ConvNets). Kd-networks perform on par with the system of Su~et~al. that is based on multi-view ConvNet\cite{Su15} and generally better than other methods in case of pose normalized dataset. For the perturbed version of the dataset, Kd-network suffer from degradation in performance due to sensitivity to global rotation. Multi-view (20 random views) version of Kd-network again perform on par with most sophisticated multi-view ConvNets.}
  \label{tab:retrieval}
\end{table*}

Finally, in \tab{class_int} we assess the importance of two different augmentations as well as the relative performance of randomized and deterministic trees. These experiments suggest that the randomization of kd-tree boosts the performance (generalization) considerably, while the geometric augmentations give a smaller effect.

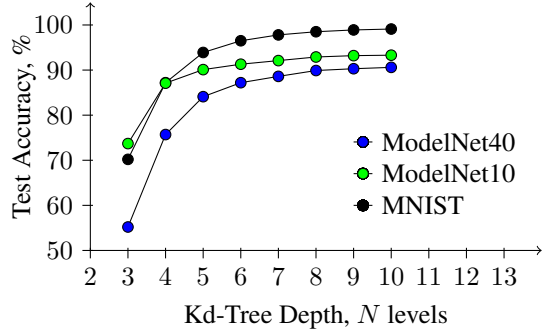
\begin{figure}
    \centering
    \begin{tikzpicture}[x=.5cm,y=.06cm]
  \draw[->,xshift=-1cm] (2,50) -- coordinate (x axis mid) (14, 50);
  \draw[->,xshift=-1cm] (2,50) -- coordinate (y axis mid) (2, 105);
  
  \foreach \x in {2,...,13}
    \draw[xshift=-1cm,yshift=3cm] (\x,1pt) -- (\x,-3pt)
      node[anchor=north] {$\x$};
  \foreach \y in {50,60,...,100}
    \draw (1pt,\y) -- (-3pt,\y)
      node[anchor=east] {$\y$};
  
  \node[below=0.6cm] at (x axis mid) {Kd-Tree Depth, $N$ levels};
  \node[rotate=90,above=0.6cm] at (y axis mid) {Test Accuracy, \%};
  
  \draw[xshift=-1cm] plot[mark=*]
    file {mnist_depth.data};
  \draw[xshift=-1cm] plot[mark=*, mark options={fill=green}]
    file {modelnet10_depth.data};
  \draw[xshift=-1cm] plot[mark=*, mark options={fill=blue}]
    file {modelnet40_depth.data};
    
  \begin{scope}[shift={(9,60)}]
    \draw[xshift=-1cm] (0,0) --
      plot[mark=*] (0.25,0) -- (0.5,0)
      node[right]{MNIST};
    \draw[xshift=-1cm,yshift=\baselineskip] (0,0) --
      plot[mark=*, mark options={fill=green}] (0.25,0) -- (0.5,0)
      node[right]{ModelNet10};
    \draw[xshift=-1cm,yshift=2\baselineskip] (0,0) --
      plot[mark=*, mark options={fill=blue}] (0.25,0) -- (0.5,0)
      node[right]{ModelNet40};
  \end{scope}
\end{tikzpicture}
    \caption{Kd-tree depth experiments. Test accuracy for Kd-networks trained on clouds of different size $2^N$ (corresponding to kd-tree depth $N$). Saturation without overfitting can be observed.}
    \label{fig:depth}
\end{figure}

\textbf{Kd-tree depth experiments.} For better understanding of the effect of depth, we also conducted a series of experiments corresponding to trees of different depths (\fig{depth}) of less or equal than ten. To obtain Kd-network architectures for smaller depths we simply remove initial layers from our 10-depth architecture (described above).  

Apart from the saturating performance, we observe that the learning time for each epoch for smaller models becomes very short but the number of epochs to achieve convergence increases. For bigger models the time of kd-tree construction (and point sampling) becomes the bottleneck in our implementation.

\textbf{Degradation in the presence of non-uniform sampling and jitter.} We have also measured the degradation of Kd-networks in the presence of non-uniform sampling and jitter and provide the results in the supplementary material. Overall, degradation from both effects on the ModelNet10 benchmark is surprisingly graceful.

\subsection{Shape retrieval}

\textbf{Dataset and data processing.} For the purpose of evaluation for 3D shape retrieval task we use ShapeNetCore dataset~\cite{Chang15}. ShapeNetCore is a subset of full ShapeNet dataset of 3D shapes with manually verified category annotations and alignment. It consists of 51300 unique 3D shapes divided into 55 categories each represented by its triangular meshes. For our experiments we used a distribution of the dataset and a training/validation/test split provided by the organizers of 3D Shape Retrieval Contest 2016 (SHREC16) \cite{Savva16}. Apart from the aligned shapes this distribution contains a perturbed version of the dataset, which consists of the same shapes each perturbed by a random rotation. Also, there is an additional division into several subcategories available for each category. In our experiments we evaluate on both versions of the dataset.

\textbf{Training and test procedures.} We used a two stage training procedure for the object retrieval task. Firstly, the network was trained to perform classification task in the manner described above. Secondly, the final layer of the network predicting the class posterior was removed, resulting representations of point clouds were normalized and used as shape descriptors provided for the fine-tuning of the network with histogram loss. A mini-batch of size $110$ was used for training, each containing two randomly selected shapes from each category of the dataset. Both training and prediction was done with geometric perturbations and kd-tree randomization applied. The parameters of the augmentations were taken from the classification task. To improve stability and quality of prediction at test time for each model the descriptors were averaged over several ($16$ in this experiment) randomized kd-trees before normalization.

\textbf{Benchmarking retrieval performance.} We compare our results \tab{retrieval} with the results of the participants of SHREC'16 for both normal and perturbed versions of ShapeNetCore. Most participating teams of SHREC'16 challenge used systems based on multi-view 2D ConvNets. We use the metrics introduced in \cite{Savva16}. Macro averaged metrics are computed by simple averaging of a metric across all shape categories, micro averaged metrics are computed by weighted averaging with weights proportionate to the number of shapes in a category. A depth-15 Kd-network trained with the histogram loss \cite{Ustinova16} was used for this task with leaf representation of size $16$ (obtained from the three coordinates using an additional multiplicative layer) and intermediate representations of sizes $32-32-64-64-128-128-256-256-512-512-1024-1024-2048-2048-512$. The obtained descriptors of size 512 were used to compute similarity and make predictions for each shape. A similarity cutoff was chosen from the results obtained on the validation part of the datasets.

In general our method performs on par with the system based on multiview CNNs \cite{Su15}, and better than other systems that participated in SHREC'16 for the `normal' set. For the `perturbed' version, the performance of Kd-networks suffers from non-invariance to global rotations. To address this, we implemented a simple modification (in the spirit of the TI-Pooling \cite{Laptev16}) that applies Kd-network (depth 10) to 20 different random rotations of a model and performs max-pooling over the produced representations followed by three fully connected layers to produce final shape descriptors. The resulting system achieved a competitive performance on the `perturbed' version of the benchmark (\tab{retrieval}).

\begin{table*}
  \begin{center}
  \tabcolsep=0.11cm
\scalebox{0.95}{
\begin{tabular}{|l|l|llllllllllllllll|}
\hline
         & {\small mean} & {\small aero} & {\small bag}  & {\small cap}  & {\small car}  & {\small chair} & {\small ear} & {\small guitar} & {\small knife} & {\small lamp} & {\small laptop} & {\small motor} & {\small mug}  & {\small pistol} & {\small rocket} & {\small skate} & {\small table} \\
         &      & {\small plane} &      &      &      &       & {\small phone} &      &       &      &        & {\small bike}  &      &        &
                & {\small board} &       \\
\hline
{\small Yi~\cite{Yi16}}         & 81.4 & 81.0 & 78.4 & 77.7 & 75.7 & 87.6  & 61.9 & 92.0 & 85.4 & 82.5 & 95.7 & 70.6 & 91.9 & 85.9 & 53.1 & 69.8 & 75.3  \\
{\small 3DCNN~\cite{Qi16b}}     & 79.4 & 75.1      & 72.8 & 73.3 & 70.0 & 87.2  & 63.5     & 88.4   & 79.6  & 74.4 & 93.9   & 58.7      & 91.8 & 76.4   & 51.2   & 65.3       & 77.1  \\
{\small PointNet~\cite{Qi16b}}  & 83.7 & 83.4      & 78.7 & 82.5 & 74.9 & 89.6  & 73.0     & 91.5   & 85.9  & 80.8 & 95.3   & 65.2      & 93.0 & 81.2   & 57.9   & 72.8       & 80.6   \\
\hline
{\small Kd-network}             & 82.3 & 80.1 & 74.6 & 74.3 & 70.3 & 88.6 & 73.5 & 90.2 & 87.2 & 81.0 & 94.9 & 57.4 & 86.7 & 78.1 & 51.8 & 69.9 & 80.3   \\
\hline
\end{tabular}}
  \end{center}
  \caption{Part segmentation results on ShapeNet-core dataset. The Intersection-over-Union scores are presented for each category as well as mean IoU are reported. Kd-network do not outperform PointNet, although for some classes the performance of Kd-networks is competitive or better.}
  \label{tab:segm}
\end{table*}

\subsection{Part Segmentation}

\begin{figure*}
    \begin{center}
    \scalebox{0.95}{
    \includegraphics[width=\textwidth]{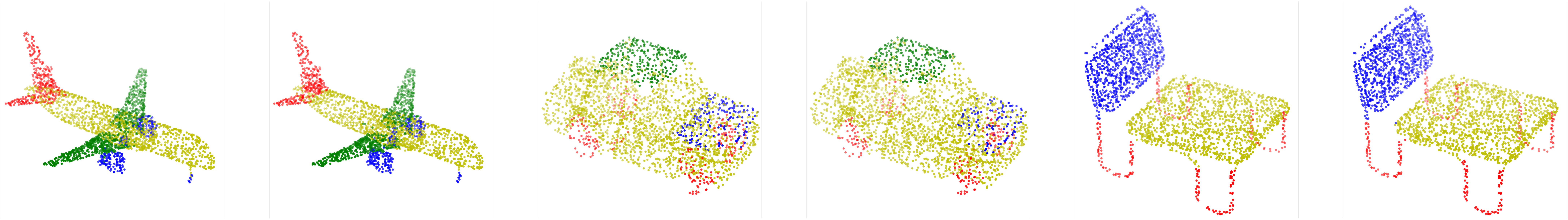}
    }
    \end{center}
    \caption{Examples of part segmentation resulting point labeling (use zoom-in for better viewing). Each pair of shapes contain ground truth labeling on the left and predicted labeling on the right. The examples were randomly taken from the validation part of the ShapeNet-core dataset.}
    \label{fig:segm_ex}
\end{figure*}

Finally, we used the architecture discussed in \sect{segm} to predict part labels for individual points within point clouds (e.g.\ in an airplane each point can correspond to body, wings, tail or engine).

\textbf{Dataset and data processing.} We evaluate our architecture for part segmentation on ShapeNet-part dataset from~\cite{Yi16}. It contains 16881 shapes represented as separate point clouds from 16 categories with per point annotation (with 50 parts in total). In this dataset, both the categories and the parts within the categories are highly imbalanced, which poses a challenge to all methods including ours.

\textbf{Training and test procedures.} Since the number of points representing each model differs in the dataset, we upsample each point cloud to size $4096$ by duplicating random points with an addition of a small noise. Apart from making data feasible for our method, such upsampling helps with rare classes. The upsampled point clouds then are fed to the architecture shown in the \fig{kdsegm}, which is optimized with the mean cross entropy over all points in a cloud as a loss function. During test time predictions are computed for the upsampled clouds, then the original cloud is passed through a constructed kd-tree to obtain a mapping of each leaf index to corresponding set of original points. This is further used to produce final predictions for every point. Similar to other tasks, we have used data augmentations both during training and test times and averaged predictions over multiple kd-trees.

\textbf{Benchmarking part segmentation performance.} Our results are compared to 3D-CNN (reproduced from \cite{Qi16b}), PointNet architecture~\cite{Qi16b}, and the architecture of~\cite{Yi16}. For each category mean intersection over union (IoU) is considered as a metric: for each shape IoUs are computed as an average of IoUs for each part which is possible to occur in this shape's category. Resulting shape IoUs are averaged over all the shapes in the category. A depth 12 variant of Kd-network was used for this task with leaf representations of size 128 and intermediate representations of sizes $128-128-128-256-256-256-256-512-512-512-512-1024$. Two additional fully connected layers of sizes $512$ and $1024$ was used in the bottleneck of the architecture. The output of segmentation network is further processed by three affine transformations interleaved with ReLU non-linearities of sizes $512$, $256$, $128$. The probabilities of the $50$ parts present in all classes in the dataset are predicted (the probabilities of the parts that are not possible for a given class are ignored following the protocol of \cite{Qi16b}). Batch-normalization is applied to each layer of the whole architecture.

The performance of Kd-networks (\tab{segm}) for the part segmentation task is competitive though not improving over state-of-the-art. We speculate that one of the reasons could be insufficient propagation of information across high-level splits within kd-tree, although resulting segmentations do not usually show the signs of underlying kd-tree structure (\fig{segm_ex}). A big advantage of Kd-networks for the segmentation task is their low memory footprint. Thus, for our particular architecture, the footprint of one example during learning is less than 120 Mb.

\section{Conclusion}
\label{sect:conclusion}

In this work we propose new deep learning architecture capable of production of representations suitable for different 3D data recognition tasks which works directly with point clouds. Our architecture has many similarities with convolutional networks, however it uses kd-tree rather than uniform grids to build the computational graphs and to share learnable parameters. With our models we achieve results comparable to current state-of-the-art for a variety of recognition problems. Compared to the top-performing convolutional architectures, kd-trees are also efficient at test-time and train-time.

The competitive performance of our deep architecture based on kd-trees suggests that other hierarchical 3D space partition structures, such as octrees, PCA-trees, bounding volume hierarchies сould be investigated as underlying structures for deep architectures.

{\bf Acknowledgement:} this work is supported by the Russian MES grant RFMEFI61516X0003.


\FloatBarrier

\ifnum\value{page}>8 \errmessage{Number of pages exceeded!!!!}\fi
{\small
\bibliographystyle{ieee}
\bibliography{refs}
}
\end{document}